\documentclass[11pt]{article}

\usepackage[final]{acl}

\usepackage{times}
\usepackage{latexsym}
\usepackage{cleveref}
\usepackage{booktabs}
\usepackage{graphicx}
\usepackage{colortbl}
\usepackage{marvosym}
\usepackage[T1]{fontenc}

\usepackage[utf8]{inputenc}

\usepackage{microtype}

\usepackage{inconsolata}

\usepackage{graphicx}
\usepackage{booktabs}        
\usepackage{xcolor}
\usepackage[table]{xcolor}   
\usepackage{multirow}
\usepackage{soul}

\sethlcolor{yellow}

\title{CRAFT: \underline{C}ritic-\underline{R}efined \underline{A}daptive Key-\underline{F}rame \underline{T}argeting for Multimodal Video Question Answering}

\author{
\textnormal{Mahesh Bhosale}$^{1}$\thanks{Equal contribution.}
\textsuperscript{\thanks{Correspondence: \texttt{mbhosale@buffalo.edu}.}}
\quad
\textnormal{Abdul Wasi}$^{1*}$
\quad
\textnormal{Vishvesh Trivedi}$^{2*}$ \\
\textnormal{Pengyu Yan}$^{1}$
\quad
\textnormal{Akhil Gorugantu}$^{1}$
\quad
\textnormal{David Doermann}$^{1}$ \\
\\
$^{1}$University at Buffalo
\quad
$^{2}$New York University
}

\begin{document}
\maketitle
\begin{abstract}
Grounded multi-video question answering over real-world news events requires systems to surface query-relevant evidence across heterogeneous video archives while attributing every claim to its supporting source. We introduce CRAFT (Critic-Refined Adaptive Key-Frame Targeting), a query-conditioned pipeline that combines dynamic keyframe selection, per-video ASR with multilingual fallback, and a hybrid critic loop to iteratively verify and repair claims before consolidation. The pipeline integrates UNLI temporal entailment, DeBERTa-v3 cross-claim screening, and a Llama-3.2-3B adjudicator, with a final citation-merging stage that emits each fact once with all supporting source identifiers. On MAGMaR~2026, CRAFT achieves the best overall average (0.739), reference recall (0.810), and citation F1 (0.635). We further evaluate on a MAGMaR-style conversion of WikiVideo with 52 non-overlapping event queries, where CRAFT also performs strongly (0.823 Avg), showing that its claim-centric evidence aggregation generalizes beyond MAGMaR. Ablations show that atomic claims, ASR, and the critic loop drive the main gains over the vanilla query-conditioned baseline. Code and implementation details are publicly available at \url{https://github.com/bhosalems/CRAFT}.


\end{abstract}

\section{Introduction}

Multi-video question answering over real-world news events underlies
tasks from event understanding to fact-checking and crisis reporting.
Recent benchmarks such as MultiVENT~2.0~\citep{kriz2025multivent2} and
the WikiVideo article-generation task~\citep{martin2025wikivideo}
formalize a strict variant of this problem: given a query and a
collection of relevant videos, a system must produce a report whose
every statement is grounded in identifiable visual, textual, or
spoken evidence from the source videos. The MAGMaR~2026 oracle task
adds two further constraints. Each query is paired with a persona
and a background paragraph, and the resulting report is scored on
six axes that separately measure content precision and recall
(\textsc{Ref-P}, \textsc{Ref-R}) against a reference answer and
citation precision and recall (\textsc{Cite-P}, \textsc{Cite-R})
against gold source videos. A high-scoring system must both surface
the right facts and attach the right videos to them.

Three properties of long news video make this hard. First,
vision-language models face a hard token-budget bottleneck on
hour-scale input: even at 1\,FPS, a long video exceeds practical
context windows~\citep{tang2025aks,gao2026apvr}, and uniform sampling
silently truncates whatever falls outside the budget. Second, even
when relevant frames are presented, recent hallucination
benchmarks~\citep{wang2024videohallucer,li2025vidhalluc,zhang2024eventhallusion}
show that VLMs routinely emit claims unsupported by the visual
content, with errors concentrated at long-tail entities, numerical
details, and event timing-precisely the content most likely to be
cited in a news report. Third, much of the answer-relevant content
in news video is spoken rather than shown: visual-only extraction
misses interview answers, on-the-ground reporting, and official
statements, especially in non-English coverage.

Prior work addresses these challenges in isolation. Adaptive
keyframe
selectors~\citep{tang2025aks,gao2026apvr}
trim the visual input to query-relevant frames but treat the result
as terminal evidence, with no check that downstream claims are
actually supported. Critic-driven video QA
systems~\citep{liu2026videomind,dang2025mupa} add verification, but
typically at the final answer-aggregation stage and at the
granularity of a single role rather than per claim. Modular
video-RAG
pipelines~\citep{jeong2025videorag,ren2025videorag,zeng2025scenerag}
compose retrieval and reasoning over long context but rely on a
single visual stream and ignore speech leading to citation faithfulness diverging
from citation correctness \cite{wallat2025faithfulness}.

\begin{figure*}
    \centering
    \includegraphics[width=0.98\linewidth]{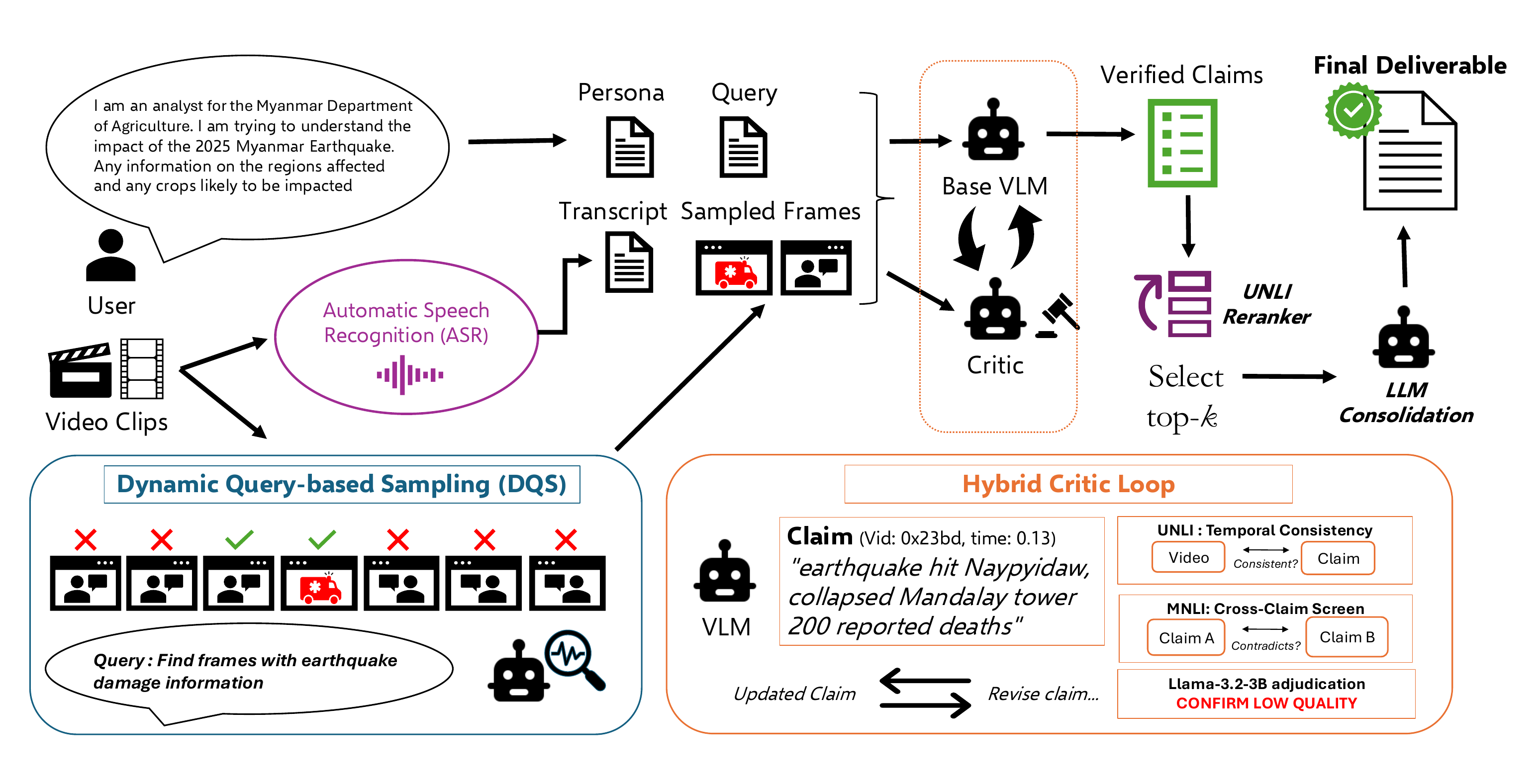}
    \caption{%
    \textbf{Overview of CRAFT.}
    Given a persona, query, and relevant videos, CRAFT builds a
    query-specific multimodal evidence stream: each video is
    transcribed once via ASR, and \emph{Dynamic Keyframe Selection}
    (DKS) selects the frames most relevant to the query. The base
    VLM consumes the persona, query, transcript, and sampled frames
    to produce atomic claims, which are refined by a \emph{hybrid
    critic loop}---UNLI for temporal grounding, MNLI for cross-claim
    contradiction screening, and a Llama-3.2-3B adjudicator that
    confirms low-quality claims and returns repair feedback for
    re-extraction. Verified claims are UNLI-reranked, and the
    top-$k$ are consolidated by an LLM into a report with every
    statement traceable to its source video and timestamp.%
  }
  \label{fig:teaser}
\end{figure*}

We present \textbf{CRAFT} (Critic-Refined Adaptive Key-Frame
Targeting), a query-conditioned pipeline that integrates these
threads for the MAGMaR~2026 oracle task (Figure~\ref{fig:teaser}).
Our contributions are:
(i) a \emph{multimodal evidence
stream}~(§\ref{sec:evidence-stream}) combining 120-second video
chunking, per-video ASR (Qwen3-ASR-1.7B with a Whisper-large-v3
fallback for low-resource languages), automatic English translation,
and dynamic query-conditioned keyframe selection, so the VLM
receives a clip and transcript both targeted at the current query;
(ii) a \emph{critic-guided extraction
loop}~(§\ref{sec:critic}) that runs a UNLI video-claim
entailment model for temporal grounding, a DeBERTa-v3 MNLI
cross-encoder for cross-claim contradiction screening, and a
Llama-3.2-3B adjudicator that confirms contradictions and emits
repair feedback, returning the critic report to the VLM for up to
four re-extraction rounds; and
(iii) atomic claim formatting (§\ref{sec:atomic-claim}) with
\emph{citation-merging}
consolidation~(§\ref{sec:packets-inference}), which emits each fact once
with all supporting source identifiers attached, preserving citation
recall while suppressing the redundancy that inflates
reference-precision loss.

On MAGMaR~2026~(§\ref{sec:experiments}), CRAFT outperforms strong
baselines with the highest overall average (0.739), reference recall
(0.810), and citation F1 (0.635) of all evaluated configurations.
Ablations~(§\ref{sec:ablations}) show that the gains from the critic
loop, atomic claims, and ASR-augmented extraction are partly
orthogonal to the choice of base VLM, transferring across
Qwen3.5-9B~\citep{qwen3.5} and Qwen3-VL-30B~\citep{bai2025qwen3vl}
backbones, and outperforming strong VLMs such as
Molmo2-8B~\citep{deitke2025molmo} and
Gemma-4-31B\footnote{\url{https://huggingface.co/google/gemma-4-31B-it}}.  
\section{Related Work}
\label{sec:related}

\paragraph{Long-video understanding with vision-language models.}
Open-source video-language models have improved rapidly along two axes: backbone capacity and temporal modeling. The Qwen-VL family progressed from dynamic-resolution and time-aligned M-RoPE in Qwen2.5-VL \citep{bai2025qwen25vl} to interleaved M-RoPE, DeepStack cross-layer fusion, and explicit timestamp tokens in Qwen3-VL \citep{bai2025qwen3vl}, while InternVL3 \citep{zhu2025internvl3} introduced Variable Visual Position Encoding and native multimodal pre-training. LLaVA-Video \citep{zhang2024llavavideo} and LLaVA-OneVision \citep{li2024llavaonevision} consolidated the LLaVA recipe for video instruction tuning. Despite these gains, all such models face a hard token-budget bottleneck on hour-scale input: even at 1\,FPS, a long video produces token counts that exceed practical context windows \citep{tang2025aks,gao2026apvr}. Specialized long-context architectures, including LongVU \citep{shen2024longvu}, Video-XL \citep{shu2025videoxl}, MovieChat \citep{song2024moviechat}, and MA-LMM \citep{he2024malmm}, mitigate this through spatiotemporal compression, sparse memory, or hierarchical attention, but typically at the cost of fine-grained temporal evidence that is essential for citation-grounded answering.

\paragraph{Adaptive keyframe selection.}
Because uniform sampling is the dominant performance bottleneck on long videos, a substantial body of recent work has focused on query-conditioned frame selection. AKS \citep{tang2025aks} formulates selection as a joint optimization over prompt-frame relevance and temporal coverage, solved by a recursive split-and-judge algorithm; APVR \citep{gao2026apvr} extends this idea to a two-granularity hierarchy in which Pivot Frame Retrieval expands the query into semantic facets and Pivot Token Retrieval performs query-aware token selection within retained frames. VideoTree \citep{wang2025videotree} replaces flat selection with a query-adaptive tree of clustered keyframes captioned coarse-to-fine. Other recent variants include MDP3 \citep{sun2025mdp3}, which casts selection as a Markov decision process; Q-Frame \citep{zhang2025qframe}, which ranks frames into multiple resolution tiers; AdaRD-Key \citep{zhang2025adardkey}, which encourages diversity through determinantal point processes; F2C \citep{sun2025f2c}, which extends keyframes to short clips to preserve motion continuity; and VidF4 \citep{liang2024vidf4}, which proposes differentiable frame scoring for end-to-end VideoQA. A.I.R.\ \citep{zou2025air} and T* \citep{ye2025tstar} replace lightweight CLIP-based scorers with iterative VLM-based reasoning over candidate frames, trading cost for accuracy. A common property of these selectors is that their output is treated as the terminal evidence representation, with no mechanism to detect whether claims subsequently extracted from the chosen frames are actually supported by the video.

\paragraph{Modular and agentic video pipelines.}
Modular pipelines decompose video question answering into captioning, retrieval, and reasoning stages. LLoVi \citep{zhang2024llovi} demonstrated that short-clip captions plus an LLM aggregator can match dedicated video models on long-form benchmarks. VideoAgent \citep{wang2024videoagent} introduced an iterative agent that uses CLIP-based frame retrieval and self-reflective stopping, achieving strong results on EgoSchema and NExT-QA with fewer than ten frames on average. MoReVQA \citep{min2024morevqa} showed that a multi-stage event-parser, grounding, and reasoning architecture with shared external memory outperforms single-stage program-generation approaches. More recent agentic systems, including VideoAgent2 \citep{zhi2025videoagent2}, Deep Video Discovery \citep{zhang2025dvd}, and VideoDeepResearch \citep{yuan2025videodr}, equip a reasoning model with multi-granular search tools over a structured video index. These systems generally place verification, when present at all, at the final answer-aggregation stage rather than during evidence extraction.

\paragraph{Critic-driven refinement and faithfulness.}
Several lines of work have explored verification and critic loops to improve grounding. In text generation, Self-RAG \citep{asai2024selfrag} and CRAG \citep{yan2024crag} introduce reflection tokens or evaluators that trigger retrieval correction. For video, VideoMind \citep{liu2026videomind} defines four explicit roles-planner, grounder, verifier, and answerer-instantiated as Chain-of-LoRA adapters, and demonstrates that the verifier role substantially improves grounding accuracy. MUPA \citep{dang2025mupa} runs three reasoning paths in parallel and consolidates them through a reflection agent. \citet{wallat2025faithfulness} further show that, in retrieval-augmented generation, citation correctness diverges sharply from citation faithfulness, motivating verification as a first-class component. Hallucination benchmarks for video, including VideoHallucer \citep{wang2024videohallucer}, EventHallusion \citep{zhang2024eventhallusion}, and VidHalluc \citep{li2025vidhalluc}, document that vision-language models routinely emit unsupported claims even when relevant frames are available. CRAFT builds on this line of work by applying a hybrid critic with iterative repair feedback at the claim level, at finer granularity than the single verifier role of \citet{liu2026videomind}.

\paragraph{Multi-video corpora and grounded generation.}
At the corpus level, MultiVENT~2.0 \citep{kriz2025multivent2} provides a large-scale multilingual benchmark of event-centric news videos, accompanied by retrieval baselines such as MMMORRF \citep{samuel2025mmmorrf} that fuse modality-specific scores via weighted reciprocal rank fusion. WikiVideo \citep{martin2025wikivideo} formalizes the task of generating articles whose every claim is grounded in audio, video, or on-screen text from a video collection. VideoRAG variants \citep{jeong2025videorag,ren2025videorag} extend retrieval-augmented generation to long-context video, while SceneRAG \citep{zeng2025scenerag} substitutes scene-level segmentation for fixed chunking. Our pipeline follows the claim-centric formulation introduced by WikiVideo and instantiates it for the multi-video setting with explicit citation merging at consolidation.

\section{Method}
\label{sec:method}

We propose a query-conditioned multimodal video question answering pipeline for
the MAGMaR~2026 oracle task, where each query is paired with a set
of relevant videos. This setting differs from standard single-video VQA because
the answer may require evidence distributed across multiple videos. Moreover, irrelevant or redundant clips can easily introduce unsupported claims. Our
pipeline, similar to \citet{martin2025wikivideo}, therefore follows a claim-centric design: it first extracts atomic,
source-grounded claims from each query-video pair, verifies them with a hybrid
critic, ranks them using video-claim support scores, and finally consolidates
them into a citation-backed report.

\subsection{Evidence Stream}\label{sec:evidence-stream}

\subsubsection{Preprocessing.}
We preprocess long source videos by splitting them into fixed-size chunks of at most 120 seconds using PyAV. This prevents the VLM from silently truncating long videos under a fixed
frame budget and allows each segment to be processed without exceeding memory
or context constraints. We retain a mapping from each chunk identifier to its
parent video identifier, and use this mapping to restore parent video IDs and consolidate the outputs.

\subsubsection{Per-video ASR and translation.}
Each unique video is transcribed once and cached for reuse. We use
Qwen3-ASR-1.7B~\citep{shi2026qwen3} as the primary ASR backend. For languages
outside its supported set in our data, such as Burmese and Nepali, we fall back
to Whisper-large-v3~\citep{radford2022whisper}. For non-English videos, we also
run a translation pass to obtain an English transcript. During claim extraction,
we provide both the original transcript and the English translation to the VLM,
allowing the model to ground claims in spoken content as well as visual
evidence.
Because ASR systems can produce repetitive token loops on low-resource or noisy
audio~\citep{koenecke2024careless}, we filter degenerate transcripts before
they reach the VLM. 
We flag a transcript as unreliable if it contains at
least 20 tokens and has very low lexical diversity, measured by a type-token
ratio below $0.18$, where the type-token ratio is the number of unique tokens
divided by the total number of tokens. We also flag transcripts with obvious
local repetition, such as the same token appearing at least 8 times
consecutively, or phrase-level repetition, where a single 3-token phrase
accounts for at least 40\% of all 3-token phrases in the transcript. Flagged
transcripts are excluded from the prompt to avoid propagating ASR artifacts into
downstream claims. Although this filtering may discard useful information from resource-scarce-language videos, stronger multilingual ASR systems could mitigate this limitation, which
we leave for future work.

\subsubsection{Dynamic Keyframe Selection.}
\label{sec:dks}

Long videos contain many frames that are irrelevant to a given query, and
uniform frame sampling can dilute the visual evidence passed to the VLM. We
therefore use Dynamic Keyframe Selection (DKS) to construct a compact visual
input for each query-video pair. DKS is applied independently for each pair
$(q,v)$, so the same video may yield different selected frames for different
queries.

For a query $q$ and video $v$, we first sample candidate frames at a fixed
temporal rate. Each frame is embedded with a visual encoder and scored against
the query embedding using image-text similarity:
\[
s_i = \mathrm{sim}\big(\phi_I(f_i), \phi_T(q)\big),
\]
where $\phi_I$ and $\phi_T$ denote the image and text encoders. The resulting
scores form a query-conditioned relevance curve over the video. We use CLIP~\cite{clip} Image and Text encoders.

We then select frames that balance high relevance with temporal coverage, similar to~\cite{tang2025aks}. The
selected frame indices are sorted in temporal order and re-encoded as a short
query-specific clip. During claim extraction, the resolver first checks whether
a DKS clip exists for the current query-video pair. If available, the VLM
receives this compact clip instead of the full chunked video; otherwise, the
pipeline falls back to the original chunk. Thus, DKS focuses visual input on
query-relevant evidence while remaining optional and non-blocking.

\subsection{Query-Conditioned Claim Extraction}
\label{sec:extraction}
Given the evidence stream for a query-video pair, we extract a set of
source-grounded claims. For each query $q$ and each video
$v \in \mathcal{V}_q$ associated with that query, we issue one VLM call to
Qwen3.5-9B~\citep{qwen3.5} served with vLLM~\citep{kwon2023efficient}. The prompt
contains the persona title, persona background, query text, the resolved video
input from the evidence stream, and the cached ASR transcript when available. If persona title/background is not available based on the query and claims we use LLM~\cite{qwen3.5} to generate it in preprocessing. The model is instructed to output \textit{atomic} claims, where each claim is a single declarative statement that can be judged as supported or unsupported by the source video. 

This produces an initial per-video claim set
$\mathcal{C}^{0}_{q,v}$ for each query-video pair. Claim extraction is
performed independently for each video so that every claim remains tied to a
specific source video, timestamp, and evidence modality.


\paragraph{Atomic claim format.} \label{sec:atomic-claim}
Each extracted claim must be independently verifiable. We discourage compound
claims that combine multiple events, entities, or causal relations into a
single sentence, since such claims become unsupported if any sub-clause is not
grounded in the video. Each claim is also tagged with its evidence modality,
such as visual evidence, on-screen text, transcript, or ASR-derived speech.

\subsection{Critic-Guided Claim Refinement}
\label{sec:critic}

The initial VLM extraction can still produce claims with weak visual grounding,
incorrect temporal references, or contradictions. To reduce these errors, we
apply a critic-guided refinement loop separately to each query-video claim set
$\mathcal{C}^{0}_{q,v}$. The loop runs for up to $R=4$ rounds and combines
three complementary critics.

The critic loop targets three distinct error types. First, a UNLI-based
video-claim entailment model~\citep{chen-etal-2020-uncertain} checks temporal grounding by scoring each claim against its cited video segment.  
Claims scoring below $0.05$ are marked as unsupported at the cited timestamp and ignored, while scores in $[0.05,0.5)$ are treated as weak support and warrant re-extraction. This filters claims that may be plausible but are not grounded in the selected temporal window. 

Second, a DeBERTa-v3 MNLI cross-encoder~\citep{He2023DeBERTaV3} screens the
per-video claim set for possible contradictions. For each pair of claim texts,
the cross-encoder estimates entailment, neutrality, and contradiction
probabilities. 
Pairs whose contradiction probability exceeds a low threshold of 0.5
are retained as candidates for further stage. We use this stage as a high-recall filter rather than a final judge, since text-only NLI can produce false positives for claims
that mention related but compatible facts. 

Third, a Llama-3.2-3B adjudicator~\citep{llama323b} verifies the candidate
contradictions. 
Given the two claims and the MNLI contradiction score, it
decides whether the claims are genuinely inconsistent spitting binary output and, if it is inconsistent, it also returns an
explanation and a repair hint. The critic report is then fed back to the
VLM together with the previous claim set, and the VLM re-extracts a revised
set of claims by removing unsupported statements, correcting weakly grounded
claims, or resolving contradictions. The loop terminates early when the claim
set no longer changes. We denote the final refined per-video claim set as
$\mathcal{C}_{q,v}$.

\subsection{Query-Level Evidence Pooling}
\label{sec:evidence_pooling}

After per-video refinement, we aggregate claims across all videos associated
with the same query. For a query $q$, the refined claims from each relevant
video $v \in \mathcal{V}_q$ are concatenated into a query-level evidence pool:
\[
\mathcal{P}_{q} = \biguplus_{v \in \mathcal{V}_q} \mathcal{C}_{q,v}.
\]
Here, $\biguplus$ denotes concatenation of claim records, not semantic
deduplication. 
Each record remains associated with its source video, timestamp,
modality, and claim identifier. This preserves provenance when the same fact is
supported by multiple videos: overlapping claims are retained as distinct
evidence items at this stage, and redundancy is resolved only during final
inference by emitting the shared fact once with all supporting citations.

\subsection{Claim Scoring and Calibration}
\label{sec:unli}

Every refined claim in the query-level evidence pool is rescored against its
source video using the same UNLI model used by the critic. This produces a
support confidence score in $[0,1]$ for each claim. We use these scores to rank
evidence rather than apply a hard threshold, since thresholding can remove rare
but useful evidence from long-tail videos.

For each query, the top-ranked claims form a compact claim packet for
downstream inference. This packet keeps the strongest supported evidence while
retaining source identifiers required for citation generation.

\subsection{Citation-Preserving Inference}
\label{sec:packets-inference}

The final inference stage uses Qwen3.5-9B in text-only mode to convert the
calibrated claim packet into report statements. The model is constrained to
use only information present in the packet and to avoid adding new entities,
numbers, dates, or causal links.

Redundant evidence is handled by citation merging: when multiple claims support
the same fact, the report states the fact once and attaches all corresponding
source identifiers. This preserves citation coverage without repeating
semantically identical statements. Final report sections are populated directly
from the generated inferences and their associated source identifiers; during
submission formatting, chunk-level video IDs are remapped to their parent video
IDs before writing the JSONL file.
\section{Experiments} \label{sec:experiments}

\begin{table*}[tp]
\centering
\scriptsize
\setlength{\tabcolsep}{4pt}
\renewcommand{\arraystretch}{1.2}
\resizebox{\textwidth}{!}{%
\begin{tabular}{lccccccc|ccccccc}
\toprule
\multirow{2}{*}{\textbf{System}} 
& \multicolumn{7}{c|}{\textbf{MAGMaR-Test}} 
& \multicolumn{7}{c}{\textbf{WikiVideo}} \\
\cmidrule(lr){2-8} \cmidrule(lr){9-15}
& \textbf{Ref-P} & \textbf{Ref-R} & \textbf{Ref-F1} 
& \textbf{Cite-P} & \textbf{Cite-R} & \textbf{Cite-F1} & \textbf{Avg}
& \textbf{Ref-P} & \textbf{Ref-R} & \textbf{Ref-F1} 
& \textbf{Cite-P} & \textbf{Cite-R} & \textbf{Cite-F1} & \textbf{Avg} \\
\midrule

Molmo2-8B 
& 0.623 & 0.541 & 0.579 & 0.498 & 0.421 & 0.457 & 0.518
& 0.641 & 0.682 & 0.661 & 0.512 & 0.598 & 0.552 & 0.607 \\
InternVL-3.5-30B-A3B 
& 0.749 & 0.688 & 0.717 & 0.645 & 0.521 & 0.576 & 0.649
& 0.802 & 0.821 & 0.811 & 0.731 & 0.689 & 0.710 & 0.761 \\

\quad (+ ASR)
& 0.761 & 0.722 & 0.741 & 0.659 & 0.551 & 0.600 & 0.672
& 0.815 & 0.848 & 0.831 & 0.743 & 0.712 & 0.727 & 0.779 \\

Gemma-4-31B 
& 0.701 & 0.658 & 0.679 & 0.589 & 0.532 & 0.559 & 0.620
& 0.721 & 0.748 & 0.734 & 0.618 & 0.630 & 0.624 & 0.679 \\

\quad (+ ASR)
& 0.712 & 0.701 & 0.706 & 0.601 & \textbf{0.561} & 0.580 & 0.644
& 0.732 & 0.778 & 0.754 & 0.629 & 0.651 & 0.640 & 0.697 \\


\midrule

CRAFT Baseline 
& 0.437 & 0.756 & 0.430 & 0.875 & 0.251 & 0.359 & 0.518
& 0.833 & 0.834 &  0.834 & 0.951 & 0.662 & 0.764 & 0.814 \\

+ Critic Loop 
& 0.491 & 0.766 & 0.480 & 0.854 & 0.259 & 0.360 & 0.535
& 0.859 & 0.845 & 0.842 & 0.953 & 0.668 & 0.773 & 0.822 \\

+ Atomic Claims 
& 0.808 & 0.762 & 0.764 & \textbf{0.944} & 0.336 & 0.426 & 0.673
& 0.940 & 0.620 & 0.735 & 0.855 & \textbf{0.858} & \textbf{0.848} & 0.809 \\

+ ASR 
& 0.760 & \textbf{0.810} & \textbf{0.783} & 0.935 & 0.512 & \textbf{0.635} & \textbf{0.739}
& 0.871 & \textbf{0.849} & \textbf{0.854} & 0.949 & 0.656 & 0.762 & 0.823 \\

\arrayrulecolor{black!20}
\specialrule{0.2pt}{0pt}{0pt}
\arrayrulecolor{black}

$\downarrow$ frames (uniform) 
& 0.775 & 0.775 & 0.769 & 0.902 & 0.503 & 0.616 & 0.723
& 0.930 & 0.640 & 0.746 & 0.845 & 0.844 & 0.830 & 0.805 \\

$\downarrow$ frames (DKS) 
& \textbf{0.822} & 0.743 & 0.772 & 0.927 & 0.453 & 0.574 & 0.715
& \textbf{0.940} & 0.832 & 0.797 & \textbf{0.966} & 0.647 & 0.761 & \textbf{0.824} \\

\bottomrule
\end{tabular}%
}
\vspace{4pt}
\caption{\textbf{Main results on MAGMaR-Test and WikiVideo}. Baseline VLMs are evaluated both with and without ASR transcript access. All rows of CRAFT baseline for MAGMaR-Test use Qwen3.5-9B and Qwen3-VL-30B-Instruct for WikiVideo as the base VLM. Best results per column are \textbf{bolded}. Avg denotes the mean of all six metrics. We use 128 uniformly sampled frames except last two rows. $\downarrow$ denotes a reduced-frame setting used to stress test uniform sampling; DKS improves this setting by selecting more query-relevant frames, especially improving precision. For MAGMaR-Test we choose 64 reduced frames and for WikiVideo we choose 32 reduced frames.}
\label{tab:main}
\end{table*}

\begin{table}[t]
\centering
\scriptsize
\setlength{\tabcolsep}{3pt}
\renewcommand{\arraystretch}{1.05}
\resizebox{\columnwidth}{!}{%
\begin{tabular}{lccc}

\toprule
\textbf{System} 
& \textbf{ROUGE-L} 
& \textbf{BERTScore} 
& \textbf{AnsRel} \\
\midrule

\multicolumn{4}{c}{\textbf{MAGMaR-Test}} \\
\midrule

InternVL-3.5-30B-A3B 
& 0.1497 & 0.0945 & 0.6382 \\

\quad (+ ASR)
& 0.1182 & 0.0964 & 0.6462 \\

Gemma-4-31B
& 0.1426 & 0.0950 & 0.5769 \\

\quad (+ ASR)
& 0.1100 & 0.1224 & 0.5799 \\

CRAFT
& \textbf{0.1839} & \textbf{0.1709} & \textbf{0.6504} \\

\midrule

\multicolumn{4}{c}{\textbf{WikiVideo}} \\
\midrule

InternVL-3.5-30B-A3B 
& 0.1241 & -0.0184 & 0.5843 \\

\quad (+ ASR)
& 0.1265 & 0.0083 & 0.6069 \\

Gemma-4-31B
& 0.1526 & 0.0634 & 0.6486 \\

\quad (+ ASR)
& 0.1360 & 0.0632 & 0.6589 \\

CRAFT
& \textbf{0.3014} & \textbf{0.2683} & \textbf{0.6664} \\

\bottomrule
\end{tabular}
}
\vspace{4pt}

\caption{
\textbf{Generation quality comparison on MAGMaR-Test and WikiVideo.}
We report ROUGE-L, BERTScore F1, and Answer Relevance (AnsRel) for baseline VLMs with and without ASR transcript access, alongside CRAFT.
}

\label{tab:generation_quality}
\end{table}

\subsection{Benchmarks}
\paragraph{MAGMaR.} We evaluate on the MAGMaR 2026 oracle task, a multi-video question answering benchmark targeting real-world news events. The data is based on subset of WikiVideo~\cite{martin2025wikivideo}. For the retrieval and RAG settings, we retrieve relevant videos from a combination of the MAGMaR data and MultiVENT2.0 test~\cite{kriz2025multivent2}.  The dataset comprises 92 source videos with average length of 1.82 mins distributed across 10 topically diverse topics including elections, natural disasters, and geopolitical events paired with 19 official evaluation queries. Each query is associated with a set of relevant videos, and the answer may require aggregating evidence distributed across multiple clips. This multi-source setting makes the benchmark challenging because models must identify relevant evidence across heterogeneous videos while avoiding unsupported claims from irrelevant or redundant content. Each generated claim should also be accompanied by a citation to the supporting evidence video.

\paragraph{WikiVideo.} We also evaluate on the original super set dataset - WikiVideo~\cite{martin2025wikivideo}, a grounded multi-video article generation benchmark built from real-world event videos linked to Wikipedia articles. The dataset is constructed from MultiVENT 1.0 and 2.0~\cite{kriz2025multivent2} videos whose events have corresponding English Wikipedia articles, and the reference articles are derived from Wikipedia lead sections. WikiVideo contains 57 event topics spanning 427 videos with average length of 1 min from 2016 to 2025, with each event paired with an expert-written Wikipedia-style article grounded in video evidence. The annotation process decomposes Wikipedia lead sentences into atomic claims, grounds each claim in supporting video, audio, or OCR evidence, and rewrites the article so that it includes only information supported by the videos. On average, each event contains 7.65 relevant videos, 51.1 grounded subclaims, and a 118-token reference article. This makes WikiVideo well suited for evaluating whether models can synthesize high-level event information across multiple videos while maintaining claim-level grounding and citations to supporting evidence.

\subsection{Evaluation Metrics}
Predictions are evaluated using both automatic and human evaluation. For automatic evaluation, we use MiRAGE~\cite{mirage}, which assesses factuality, information coverage, groundedness, and the correctness of citation attribution. Each MiRAGE entailment judgment is judged by Qwen-7B or CLUE~\cite{clue}. Reported results in the main text use Qwen-7B, which was used during the development of our CRAFT system for submission. The official MAGMaR leaderboard uses CLUE for evaluation, we report these results in the supplementary material. For human evaluation, three annotators assign scalar scores from 1 to 5 to each system output, assessing factuality, adequacy, coherence, relevance, and fluency. After scoring all predictions, the annotators also select the best system response for each query. We report the human evaluation results in the supplementary material.

Concretely, we report six MiRAGE~\cite{mirage} metrics that evaluate both information quality and citation fidelity at the subclaim level. \textit{Reference Precision (Ref-P)} measures the proportion of generated subclaims that are supported by the reference, capturing whether the prediction contains factual and relevant information. \textit{Reference Recall (Ref-R)} measures the proportion of reference subclaims that are covered by the generated report, capturing information completeness. Their harmonic mean gives \textit{Reference F1 (Ref-F1)}. For citation evaluation, \textit{Citation Precision (Cite-P)} measures whether generated subclaims are supported by their cited source videos, while \textit{Citation Recall (Cite-R)} measures whether reference subclaims that are covered by the prediction are attributed to the correct supporting videos. Their harmonic mean gives \textit{Citation F1 (Cite-F1)}. The overall \textit{Macro-Average} is computed as the mean of the six reported metrics. Additionally, we report three complementary metrics designed to capture failure modes not explicitly measured by MiRAGE:

\textit{ROUGE-L}~\cite{lin2004rouge}, computed over the concatenated report text without stemming. Since the benchmark spans multiple languages (e.g., English, Mandarin, Burmese, and Nepali), language-specific stemming introduces substantial noise. We therefore use ROUGE-L primarily as a lightweight regression signal for lexical overlap with the reference report.

\textit{BERTScore}~\cite{zhang2020bertscore} F1 using \texttt{bert-base-multilingual-cased} with \texttt{rescale\_with\_baseline=True}. This metric captures document-level semantic similarity and stylistic alignment, complementing MiRAGE's claim-level decomposition.

\textit{RAGAS Answer Relevance}~\cite{es2024ragas}, which directly evaluates whether the persona-grounded query was meaningfully answered. For each generated report, we sample $K=3$ hypothetical questions using \texttt{Qwen2.5-7B-Instruct} (temperature $0.7$, top-$p=0.9$), embed both the reconstructed and gold queries using \texttt{Qwen3-Embedding-0.6B}, and report the mean cosine similarity.

ROUGE-L and BERTScore are reference-dependent metrics and are therefore computed only on the subsets containing gold reports (8/19 queries for MagMaR and 52/56 queries for WikiVideo). The remaining 15 queries are excluded from these metrics and explicitly marked in the results table. In contrast, Answer Relevance is reference-free and is reported for all queries.

\subsection{Baselines and Setup}
\paragraph{CRAFT Baseline.} 
We construct the CRAFT baseline as a basic pipeline for generating answers and citations given a video and its corresponding query. Additional proposed improvements are built on top of this baseline. The pipeline uses a multimodal LLM (base VLM) as the backbone: for each query, the model receives sampled video frames and is prompted to generate claims along with their supporting video citations. The model only has access to frames that are uniformly sampled from the input video, with a maximum of 128 frames provided. In the baseline, we also use UNLI~\cite{chen-etal-2020-uncertain} to re-rank the generated claims so that the downstream LLM can better prioritize important evidence. Finally, a text-only LLM aggregates the claims, removes duplicates, and consolidates them into the final response for each query. CRAFT uses base VLM as Qwen-3.5-9B-VL as a backbone for MAGMaR-Test benchmark and Qwen3-VL-30B-A3B-Instruct for Wikivideo benchmark, unless otherwise explicitly specified. For final LLM Consolidator we use Qwen3.5-9B in text-only mode. Every other addition over this baseline is described in~\cref{sec:method} and evaluated in~\cref{tab:main}. Results for CRAFT are obtained using 8 NVIDIA A6000 GPUs, and it takes $~2$ hours to get final results for Wikivideo and  $~0.75$ hour on MAGMaR-Test dataset.  
\paragraph{Other Baselines.}
We additionally evaluate a diverse set of publicly available multimodal LLMs spanning multiple architectural families and parameter scales, including Molmo2-8B~\cite{clark2026molmo2}, InternVL3-30B-A3B~\cite{zhu2025internvl3}, Qwen3-VL-30B-A3B-Instruct~\cite{bai2025qwen3vl}, and Gemma-4-31B~\cite{team2024gemma}. These comparisons provide a broader characterization of the proposed task beyond the CRAFT pipeline itself.

For all baselines, videos are represented using uniformly sampled frames. For InternVL3-30B-A3B and Gemma-4-31B, we further evaluate both \textit{visual-only} and \textit{visual+ASR} variants using the same ASR backend employed by CRAFT. Concretely, for each $(q,v)$ pair, we issue a single VLM call requesting factual claims relevant to the query and concatenate the resulting per-video generations into a final per-query report without any additional scoring, reranking, deduplication, or calibration.

Long videos are pre-segmented offline into 60-second chunks. Each chunk is sampled at 1 fps with a maximum of 60 frames per call, and generation is capped at 1024 new tokens.

In the \textit{visual+ASR} setting, we augment the visual inputs with \texttt{Whisper-large-v3} transcripts sourced from the \texttt{akhilvssg/magmar-2026-test-asr-embeddings} release on MagMaR and the corresponding WikiVideo dump. For each chunk, we provide both the original-language transcript and its English translation as auxiliary textual context.


\subsection{Main Results}
\Cref{tab:main} reports the main results on MAGMaR-Test and WikiVideo. Overall, CRAFT achieves the best average performance on MAGMaR-Test and competitive performance on WikiVideo, showing consistent gains over publicly available VLM baselines. Among the baseline models, adding ASR generally improves performance, especially on WikiVideo, indicating that explicit speech transcripts provide useful evidence beyond visual frames alone.

Within CRAFT, the largest improvement comes from moving beyond the initial baseline toward atomic claim generation and ASR-augmented evidence extraction. On MAGMaR-Test, adding atomic claims substantially improves Ref-P and Ref-F1 as compared to baseline, suggesting that decomposing evidence into finer-grained claims helps the model produce more precise and verifiable answers. Adding ASR further improves Ref-R and Cite-F1, showing that spoken content is important for recovering missing information and assigning better citations. However, citation recall remains more challenging than citation precision, indicating that exact claim-to-video attribution is still a difficult part of the task.

The last two rows simulate low-frame settings to stress test the robustness of frame sampling when only small compute budget is alloted to the task. This becomes more challenging for longer videos, where relevant information is often sparse and distributed across distant segments, making it harder to preserve context. This is reflected in the larger performance drop across most metrics on MAGMaR-Test compared to WikiVideo, as MAGMaR-Test videos are on average roughly twice as long. The $\downarrow$ frames rows denote reduced-frame settings, where fewer frames are passed to the system. In the uniform setting, the reduced frame budget is sampled uniformly, which can miss query-relevant evidence. In the DKS setting, uniform sampling is replaced with dynamic keyframe selection under the same reduced-frame budget. DKS improves precision in several cases by selecting more relevant frames, although it can trade off recall when some supporting evidence is filtered out. This suggests that adaptive frame selection is useful under constrained visual budgets, but further work is needed to balance precision-oriented keyframe selection with broad evidence coverage.

\Cref{tab:generation_quality} reports auxiliary generation-quality metrics on MAGMaR-Test and WikiVideo. CRAFT achieves the best ROUGE-L, BERTScore F1, and Answer Relevance on both datasets, indicating that its claim-centric aggregation improves not only factual grounding and citation quality, but also the fluency and relevance of the generated reports. For the baseline VLMs, adding ASR generally improves answer relevance and semantic similarity in some cases, but the gains are not consistent across all metrics.

\begin{table}[t]
\centering
\tiny
\setlength{\tabcolsep}{2.2pt}
\renewcommand{\arraystretch}{1.12}
\resizebox{\columnwidth}{!}{%
\begin{tabular}{lccccccc}
\toprule
\textbf{Variant} 
& \textbf{Ref-P} & \textbf{Ref-R} & \textbf{Ref-F1} 
& \textbf{Cite-P} & \textbf{Cite-R} & \textbf{Cite-F1} 
& \textbf{Avg} \\
\midrule

Qwen3.5-9B-VL backbone
& 0.760 & 0.810 & 0.783 
& 0.935 & 0.512 & 0.635 
& 0.739 \\

Qwen3-Omni-30B-A3B
& 0.745 & 0.761 & 0.735 
& 0.878 & 0.346 & 0.471 
& 0.656 \\

\bottomrule
\end{tabular}%
}
\vspace{2pt}
\caption{\textbf{Backbone replacement} ablation on MAGMaR-Test. Qwen3-Omni-30B-A3B directly uses audio input, while Qwen3.5-9B-VL uses ASR transcripts. Avg denotes the mean of all six metrics.}
\label{tab:backbone_ablation}
\end{table}

















\subsection{Ablation Studies} \label{sec:ablations}
\paragraph{Omni-Model.} Although Qwen3-Omni-30B-A3B directly processes audio, it does not outperform the ASR-based Qwen3.5-9B-VL backbone as seen in~\cref{tab:backbone_ablation}. This suggests that, for claim-centric video QA, explicit ASR transcripts provide a more reliable intermediate representation for evidence extraction, citation assignment, and downstream text-based verification. Direct audio conditioning may encode speech information implicitly, but it can make fine-grained details such as named entities, dates, and numerical facts harder to recover and verify. In contrast, ASR converts speech into explicit textual evidence, which better aligns with the claim aggregation and citation modules in CRAFT.

\begin{table}[t]
\centering
\tiny
\setlength{\tabcolsep}{2.2pt}
\renewcommand{\arraystretch}{1.12}
\resizebox{\columnwidth}{!}{%
\begin{tabular}{lccccccc}
\toprule
\textbf{System} & \textbf{Ref-P} & \textbf{Ref-R} & \textbf{Ref-F1} & \textbf{Cite-P} & \textbf{Cite-R} & \textbf{Cite-F1} & \textbf{Avg} \\
\midrule
CRAFT (full)                                  & 0.760 & 0.810 & 0.783 & 0.935 & 0.512 & 0.635 & 0.739 \\
\quad w/ Qwen replaces UNLI                   & 0.732 & 0.788 & 0.759 & 0.874 & 0.469 & 0.601 & 0.704 \\
\quad w/ Qwen replaces Llama-3.2-3B           & 0.763 & 0.812 & 0.787 & 0.937 & 0.516 & 0.619 & 0.732 \\
\quad w/ Qwen unified critic (no MNLI screen) & 0.743 & 0.798 & 0.770 & 0.909 & 0.493 & 0.619 & 0.722 \\
\bottomrule
\end{tabular}%
}
\caption{
\textbf{Component ablations on MAGMaR-Test.}
Replacing specialized critic components with a unified Qwen-based adjudicator consistently degrades attribution performance. The unified critic variant removes the DeBERTa-v3 MNLI screening stage and performs contradiction detection and adjudication in a single pass. We report precision (P), recall (R), and F1 for both Reference Attribution and Citation Attribution.}
\label{tab:ablations}
\end{table}
 
\paragraph{UNLI Scorer.}
Replacing UNLI with zero-shot Qwen3.5-9B causes the largest drop in citation metrics, confirming that UNLI's specialized temporal entailment training is not recoverable by a general-purpose VLM.

\paragraph{Critic Adjudicator.}
Replacing Llama-3.2-3B with Qwen3.5-9B yields a marginal drop, suggesting the 3B adjudicator is already sufficient for binary contradiction confirmation and the larger model provides no measurable benefit.

\paragraph{Unified Qwen Critic.}
Removing the DeBERTa MNLI pre-filter and collapsing screening and adjudication into a single Qwen pass degrades citation precision, showing the specialized NLI screener provides a signal that general-purpose prompting does not fully replicate.


\section{Conclusion and Future Work}

We presented CRAFT, a claim-centric pipeline for grounded multi-video question answering that combines keyframe selection, ASR-based evidence extraction, critic-guided verification, and citation-backed report generation. CRAFT improves over the baseline through atomic-claim formatting, ASR, and the critic loop. However, recall and citation recall remain challenging, suggesting that future work should improve evidence coverage, cross-video retrieval, multilingual ASR, and precise claim-to-video attribution.

\bibliography{custom}

@inproceedings{koenecke2024careless,
  title={Careless whisper: Speech-to-text hallucination harms},
  author={Koenecke, Allison and Choi, Anna Seo Gyeong and Mei, Katelyn X and Schellmann, Hilke and Sloane, Mona},
  booktitle={Proceedings of the 2024 ACM conference on fairness, accountability, and transparency},
  pages={1672--1681},
  year={2024}
}

@article{bai2025qwen25vl,
  title   = {Qwen2.5-{VL} Technical Report},
  author  = {Bai, Shuai and Chen, Keqin and Liu, Xuejing and Wang, Jialin and Ge, Wenbin and Song, Sibo and Dang, Kai and Wang, Peng and Wang, Shijie and Tang, Jun and Zhong, Humen and Zhu, Yuanzhi and Yang, Mingkun and Li, Zhaohai and Wan, Jianqiang and Wang, Pengfei and Ding, Wei and Fu, Zheren and Xu, Yiheng and Ye, Jiabo and Zhang, Xi and Xie, Tianbao and Cheng, Zesen and Zhang, Hang and Yang, Zhibo and Xu, Haiyang and Lin, Junyang},
  journal = {arXiv preprint arXiv:2502.13923},
  year    = {2025}
}

@article{bai2025qwen3vl,
  title   = {Qwen3-{VL} Technical Report},
  author  = {Bai, Shuai and Cai, Yuxuan and Chen, Ruizhe and Chen, Keqin and Chen, Xionghui and Cheng, Zesen and Deng, Lianghao and Ding, Wei and Gao, Chang and Ge, Chunjiang and Ge, Wenbin and Guo, Zhifang and Huang, Qidong and Huang, Jie and Huang, Fei and Hui, Binyuan and Jiang, Shutong and Li, Zhaohai and Li, Mingsheng and Li, Mei and Li, Kaixin and Lin, Zicheng and Lin, Junyang and others},
  journal = {arXiv preprint arXiv:2511.21631},
  year    = {2025}
}

@article{zhu2025internvl3,
  title   = {{InternVL3}: Exploring Advanced Training and Test-Time Recipes for Open-Source Multimodal Models},
  author  = {Zhu, Jinguo and Wang, Weiyun and Chen, Zhe and Liu, Zhaoyang and Ye, Shenglong and Gu, Lixin and Tian, Hao and Duan, Yuchen and Su, Weijie and Shao, Jie and Gao, Zhangwei and Cui, Erfei and Cao, Yue and Liu, Yangzhou and Wei, Xingguang and Zhang, Hongjie and Wang, Haomin and Xu, Weiye and Li, Hao and Wang, Jiahao and others},
  journal = {arXiv preprint arXiv:2504.10479},
  year    = {2025}
}

@misc{qwen3.5,
    title  = {{Qwen3.5}: Towards Native Multimodal Agents},
    author = {{Qwen Team}},
    month  = {February},
    year   = {2026},
    url    = {https://qwen.ai/blog?id=qwen3.5}
}

@misc{radford2022whisper,
  doi = {10.48550/ARXIV.2212.04356},
  url = {https://arxiv.org/abs/2212.04356},
  author = {Radford, Alec and Kim, Jong Wook and Xu, Tao and Brockman, Greg and McLeavey, Christine and Sutskever, Ilya},
  title = {Robust Speech Recognition via Large-Scale Weak Supervision},
  publisher = {arXiv},
  year = {2022},
  copyright = {arXiv.org perpetual, non-exclusive license}
}

@article{zhang2024llavavideo,
  title   = {Video Instruction Tuning with Synthetic Data},
  author  = {Zhang, Yuanhan and Wu, Jinming and Li, Wei and Li, Bo and Ma, Zejun and Liu, Ziwei and Li, Chunyuan},
  journal = {arXiv preprint arXiv:2410.02713},
  year    = {2024}
}

@article{li2024llavaonevision,
  title   = {{LLaVA-OneVision}: Easy Visual Task Transfer},
  author  = {Li, Bo and Zhang, Yuanhan and Guo, Dong and Zhang, Renrui and Li, Feng and Zhang, Hao and Zhang, Kaichen and Li, Yanwei and Liu, Ziwei and Li, Chunyuan},
  journal = {arXiv preprint arXiv:2408.03326},
  year    = {2024}
}

@article{martin2025wikivideo,
  title={Wikivideo: Article generation from multiple videos},
  author={Martin, Alexander and Kriz, Reno and Walden, William Gantt and Sanders, Kate and Recknor, Hannah and Yang, Eugene and Ferraro, Francis and Van Durme, Benjamin},
  journal={arXiv preprint arXiv:2504.00939},
  year={2025}
}

@article{shen2024longvu,
  title   = {{LongVU}: Spatiotemporal Adaptive Compression for Long Video-Language Understanding},
  author  = {Shen, Xiaoqian and Xiong, Yunyang and Zhao, Changsheng and Wu, Lemeng and Chen, Jun and Zhu, Chenchen and Liu, Zechun and Xiao, Fanyi and Varadarajan, Balakrishnan and Bordes, Florian and Liu, Zhuang and Xu, Hu and Kim, Hyunwoo J. and Soran, Bilge and Krishnamoorthi, Raghuraman and Elhoseiny, Mohamed and Chandra, Vikas},
  journal = {arXiv preprint arXiv:2410.17434},
  year    = {2024}
}

@inproceedings{kwon2023efficient,
  title={Efficient Memory Management for Large Language Model Serving with PagedAttention},
  author={Woosuk Kwon and Zhuohan Li and Siyuan Zhuang and Ying Sheng and Lianmin Zheng and Cody Hao Yu and Joseph E. Gonzalez and Hao Zhang and Ion Stoica},
  booktitle={Proceedings of the ACM SIGOPS 29th Symposium on Operating Systems Principles},
  year={2023}
}

@article{shu2025videoxl,
  title   = {{Video-XL}: Extra-Long Vision Language Model for Hour-Scale Video Understanding},
  author  = {Shu, Yan and Zhang, Peitian and Liu, Zheng and Qin, Minghao and Zhou, Junjie and Liang, Zhengyang and Huang, Tiejun and Zhao, Bo},
  journal = {arXiv preprint arXiv:2409.14485},
  year    = {2025}
}

@inproceedings{song2024moviechat,
  title     = {{MovieChat}: From Dense Token to Sparse Memory for Long Video Understanding},
  author    = {Song, Enxin and Chai, Wenhao and Wang, Guanhong and Zhang, Yucheng and Zhou, Haoyang and Wu, Feiyang and Chi, Haozhe and Guo, Xun and Ye, Tian and Zhang, Yanting and Lu, Yan and Hwang, Jenq-Neng and Wang, Gaoang},
  booktitle = {Proceedings of the IEEE/CVF Conference on Computer Vision and Pattern Recognition (CVPR)},
  year      = {2024},
  pages     = {18221--18232}
}

@inproceedings{he2024malmm,
  title     = {{MA-LMM}: Memory-Augmented Large Multimodal Model for Long-Term Video Understanding},
  author    = {He, Bo and Li, Hengduo and Jang, Young Kyun and Jia, Menglin and Cao, Xuefei and Shah, Ashish and Shrivastava, Abhinav and Lim, Ser-Nam},
  booktitle = {Proceedings of the IEEE/CVF Conference on Computer Vision and Pattern Recognition (CVPR)},
  year      = {2024}
}

@article{shi2026qwen3,
  title={Qwen3-ASR Technical Report},
  author={Shi, Xian and Wang, Xiong and Guo, Zhifang and Wang, Yongqi and Zhang, Pei and Zhang, Xinyu and Guo, Zishan and Hao, Hongkun and Xi, Yu and Yang, Baosong and others},
  journal={arXiv preprint arXiv:2601.21337},
  year={2026}
}

@inproceedings{tang2025aks,
  title     = {Adaptive Keyframe Sampling for Long Video Understanding},
  author    = {Tang, Xi and Qiu, Jihao and Xie, Lingxi and Tian, Yunjie and Jiao, Jianbin and Ye, Qixiang},
  booktitle = {Proceedings of the IEEE/CVF Conference on Computer Vision and Pattern Recognition (CVPR)},
  year      = {2025},
  pages     = {29118--29128}
}

@article{gao2026apvr,
  title   = {{APVR}: Hour-Level Long Video Understanding with Adaptive Pivot Visual Information Retrieval},
  author  = {Gao, Hong and Wang, Yiming and Hu, Xin and Cao, Xun and Tao, Mingkui},
  journal = {arXiv preprint arXiv:2506.04953},
  year    = {2025},
  note    = {To appear in AAAI 2026}
}

@inproceedings{chen-etal-2020-uncertain,
    title = "Uncertain Natural Language Inference",
    author = "Chen, Tongfei  and
      Jiang, Zhengping  and
      Poliak, Adam  and
      Sakaguchi, Keisuke  and
      Van Durme, Benjamin",
    booktitle = "Proceedings of the 58th Annual Meeting of the Association for Computational Linguistics",
    month = jul,
    year = "2020",
    address = "Online",
    publisher = "Association for Computational Linguistics",
    url = "https://aclanthology.org",
    doi = "10.18653/v1/2020.acl-main.774",
    pages = "8772--8779",
}

@inproceedings{wang2025videotree,
  title     = {{VideoTree}: Adaptive Tree-Based Video Representation for {LLM} Reasoning on Long Videos},
  author    = {Wang, Ziyang and Yu, Shoubin and Stengel-Eskin, Elias and Yoon, Jaehong and Cheng, Feng and Bertasius, Gedas and Bansal, Mohit},
  booktitle = {Proceedings of the IEEE/CVF Conference on Computer Vision and Pattern Recognition (CVPR)},
  year      = {2025},
  pages     = {3272--3283}
}

@article{sun2025mdp3,
  title   = {{MDP$^3$}: A Training-Free Approach for List-Wise Frame Selection in Video-{LLMs}},
  author  = {Sun, Hui and Lu, Shiyin and Wang, Huanyu and Chen, Qing-Guo and Xu, Zhao and Luo, Weihua and Zhang, Kaifu and Li, Ming},
  journal = {arXiv preprint arXiv:2501.02885},
  year    = {2025},
  note    = {Published at ICCV 2025}
}

@article{zhang2025qframe,
  title   = {{Q-Frame}: Query-Aware Frame Selection and Multi-Resolution Adaptation for Video-{LLMs}},
  author  = {Zhang, Shaojie and Yang, Jiahui and Yin, Jianqin and Luo, Zhenbo and Luan, Jian},
  journal = {arXiv preprint arXiv:2506.22139},
  year    = {2025},
  note    = {Published at ICCV 2025}
}

@inproceedings{He2023DeBERTaV3,
  title={De{BERT}a{V}3: Improving {DeBERT}a using {ELECTRA}-Style Pre-training with Gradient-Disentangled Embedding Sharing},
  author={Pengcheng He and Jianfeng Gao and Weizhu Chen},
  year={2023},
  booktitle={International Conference on Learning Representations ({ICLR})}
}

@article{zhang2025adardkey,
  title   = {{AdaRD-Key}: Adaptive Relevance--Diversity Keyframe Sampling for Long-Form Video Understanding},
  author  = {Zhang, Xian and Wu, Zexi and Li, Zinuo and Xu, Hongming and Gong, Luqi and Boussaid, Farid and Werghi, Naoufel and Bennamoun, Mohammed},
  journal = {arXiv preprint arXiv:2510.02778},
  year    = {2025}
}

@inproceedings{zhang2020bertscore,
  title={BERTScore: Evaluating Text Generation with BERT},
  author={Zhang, Tianyi and Kishore, Varsha and Wu, Felix and Weinberger, Kilian Q. and Artzi, Yoav},
  booktitle={International Conference on Learning Representations (ICLR)},
  year={2020}
}

@inproceedings{lin2004rouge,
  title={ROUGE: A Package for Automatic Evaluation of Summaries},
  author={Lin, Chin-Yew},
  booktitle={Text Summarization Branches Out: Proceedings of the ACL-04 Workshop},
  pages={74--81},
  year={2004}
}

@article{es2024ragas,
  title={RAGAs: Automated Evaluation of Retrieval Augmented Generation},
  author={Es, Shahul and James, Jithin and Espinosa-Anke, Luis and Schockaert, Steven},
  journal={arXiv preprint arXiv:2309.15217},
  year={2024}
}

@article{clue,
  title={Unified Multimodal Uncertain Inference},
  author={Zhang, Dengjia and Martin, Alexander and Jurayj, William and Murray, Kenton and Van Durme, Benjamin and Kriz, Reno},
  journal={arXiv preprint arXiv:2604.08701},
  year={2026}
}

@article{mirage,
  title={Seeing Through the MiRAGE: Evaluating Multimodal Retrieval Augmented Generation},
  author={Martin, Alexander and Walden, William and Kriz, Reno and Zhang, Dengjia and Sanders, Kate and Yang, Eugene and Jin, Chihsheng and Van Durme, Benjamin},
  journal={arXiv preprint arXiv:2510.24870},
  year={2025}
}

@inproceedings{zhang2024llovi,
  title     = {A Simple {LLM} Framework for Long-Range Video Question-Answering},
  author    = {Zhang, Ce and Lu, Taixi and Islam, Md Mohaiminul and Wang, Ziyang and Yu, Shoubin and Bansal, Mohit and Bertasius, Gedas},
  booktitle = {Proceedings of the 2024 Conference on Empirical Methods in Natural Language Processing (EMNLP)},
  year      = {2024},
  pages     = {21715--21737}
}

@inproceedings{wang2024videoagent,
  title     = {{VideoAgent}: Long-Form Video Understanding with Large Language Model as Agent},
  author    = {Wang, Xiaohan and Zhang, Yuhui and Zohar, Orr and Yeung-Levy, Serena},
  booktitle = {Proceedings of the European Conference on Computer Vision (ECCV)},
  year      = {2024}
}

@inproceedings{min2024morevqa,
  title     = {{MoReVQA}: Exploring Modular Reasoning Models for Video Question Answering},
  author    = {Min, Juhong and Buch, Shyamal and Nagrani, Arsha and Cho, Minsu and Schmid, Cordelia},
  booktitle = {Proceedings of the IEEE/CVF Conference on Computer Vision and Pattern Recognition (CVPR)},
  year      = {2024},
  pages     = {13235--13245}
}

@article{zhi2025videoagent2,
  title   = {{VideoAgent2}: Enhancing the {LLM}-Based Agent System for Long-Form Video Understanding by Uncertainty-Aware {CoT}},
  author  = {Zhi, Zhuo and Wu, Qiangqiang and Shen, Minghe and Li, Wenbo and Li, Yinchuan and Shao, Kun and Zhou, Kaiwen},
  journal = {arXiv preprint arXiv:2504.04471},
  year    = {2025}
}

@article{zhang2025dvd,
  title   = {Deep Video Discovery: Agentic Search with Tool Use for Long-Form Video Understanding},
  author  = {Zhang, Xiaoyi and Jia, Zhaoyang and Guo, Zongyu and Li, Jiahao and Li, Bin and Li, Houqiang and Lu, Yan},
  journal = {arXiv preprint arXiv:2505.18079},
  year    = {2025}
}

@article{yuan2025videodr,
  title   = {{VideoDeepResearch}: Long Video Understanding with Agentic Tool Using},
  author  = {Yuan, Huaying and Liu, Zheng and Zhou, Junjie and Qian, Hongjin and Wen, Ji-Rong and Dou, Zhicheng},
  journal = {arXiv preprint arXiv:2506.10821},
  year    = {2025}
}

@article{llama323b,
  title={Llama 3.2: 1B and 3B Instruct Model Card},
  author={{Meta AI}},
  year={2024},
  url={https://huggingface.co/meta-llama/Llama-3.2-3B-Instruct}
}

@inproceedings{asai2024selfrag,
  title     = {Self-{RAG}: Learning to Retrieve, Generate, and Critique through Self-Reflection},
  author    = {Asai, Akari and Wu, Zeqiu and Wang, Yizhong and Sil, Avirup and Hajishirzi, Hannaneh},
  booktitle = {The Twelfth International Conference on Learning Representations (ICLR)},
  year      = {2024}
}

@article{yan2024crag,
  title   = {Corrective Retrieval Augmented Generation},
  author  = {Yan, Shi-Qi and Gu, Jia-Chen and Zhu, Yun and Ling, Zhen-Hua},
  journal = {arXiv preprint arXiv:2401.15884},
  year    = {2024}
}

@inproceedings{liu2026videomind,
  title     = {{VideoMind}: A Chain-of-{LoRA} Agent for Temporal-Grounded Video Reasoning},
  author    = {Liu, Ye and Lin, Kevin Qinghong and Chen, Chang Wen and Shou, Mike Zheng},
  booktitle = {The Fourteenth International Conference on Learning Representations (ICLR)},
  year      = {2026}
}

@article{dang2025mupa,
  title   = {{MUPA}: Towards Multi-Path Agentic Reasoning for Grounded Video Question Answering},
  author  = {Dang, Jisheng and Song, Huilin and Xiao, Junbin and Wang, Bimei and Peng, Han and Li, Haoxuan and Yang, Xun and Wang, Meng and Chua, Tat-Seng},
  journal = {arXiv preprint arXiv:2506.18071},
  year    = {2025}
}

@inproceedings{wallat2025faithfulness,
  title     = {Correctness is not Faithfulness in Retrieval Augmented Generation Attributions},
  author    = {Wallat, Jonas and Heuss, Maria and de Rijke, Maarten and Anand, Avishek},
  booktitle = {Proceedings of the 2025 International ACM SIGIR Conference on Innovative Concepts and Theories in Information Retrieval (ICTIR)},
  year      = {2025},
  pages     = {22--32}
}

@article{clark2026molmo2,
  title={Molmo2: Open Weights and Data for Vision-Language Models with Video Understanding and Grounding},
  author={Clark, Christopher and Zhang, Jieyu and Ma, Zixian and Park, Jae Sung and Salehi, Mohammadreza and Tripathi, Rohun and Lee, Sangho and Ren, Zhongzheng and Kim, Chris Dongjoo and Yang, Yinuo and others},
  journal={arXiv preprint arXiv:2601.10611},
  year={2026}
}

@article{team2024gemma,
  title={Gemma: Open models based on gemini research and technology},
  author={Team, Gemma and Mesnard, Thomas and Hardin, Cassidy and Dadashi, Robert and Bhupatiraju, Surya and Pathak, Shreya and Sifre, Laurent and Rivi{\`e}re, Morgane and Kale, Mihir Sanjay and Love, Juliette and others},
  journal={arXiv preprint arXiv:2403.08295},
  year={2024}
}

@inproceedings{clip,
  title={Learning transferable visual models from natural language supervision},
  author={Radford, Alec and Kim, Jong Wook and Hallacy, Chris and Ramesh, Aditya and Goh, Gabriel and Agarwal, Sandhini and Sastry, Girish and Askell, Amanda and Mishkin, Pamela and Clark, Jack and others},
  booktitle={International conference on machine learning},
  pages={8748--8763},
  year={2021},
  organization={PmLR}
}

@inproceedings{deitke2025molmo,
  title={Molmo and pixmo: Open weights and open data for state-of-the-art vision-language models},
  author={Deitke, Matt and Clark, Christopher and Lee, Sangho and Tripathi, Rohun and Yang, Yue and Park, Jae Sung and Salehi, Mohammadreza and Muennighoff, Niklas and Lo, Kyle and Soldaini, Luca and others},
  booktitle={Proceedings of the Computer Vision and Pattern Recognition Conference},
  pages={91--104},
  year={2025}
}

@article{wang2024videohallucer,
  title   = {{VideoHallucer}: Evaluating Intrinsic and Extrinsic Hallucinations in Large Video-Language Models},
  author  = {Wang, Yuxuan and Wang, Yueqian and Zhao, Dongyan and Xie, Cihang and Zheng, Zilong},
  journal = {arXiv preprint arXiv:2406.16338},
  year    = {2024}
}

@article{zhang2024eventhallusion,
  title   = {{EventHallusion}: Diagnosing Event Hallucinations in Video {LLMs}},
  author  = {Zhang, Jiacheng and Jiao, Yang and Chen, Shaoxiang and Zhao, Na and Tan, Zhiyu and Li, Hao and Ma, Xingjun and Chen, Jingjing},
  journal = {arXiv preprint arXiv:2409.16597},
  year    = {2024}
}

@inproceedings{li2025vidhalluc,
  title     = {{VidHalluc}: Evaluating Temporal Hallucinations in Multimodal Large Language Models for Video Understanding},
  author    = {Li, Chaoyu and Im, Eun Woo and Fazli, Pooyan},
  booktitle = {Proceedings of the IEEE/CVF Conference on Computer Vision and Pattern Recognition (CVPR)},
  year      = {2025},
  pages     = {13723--13733}
}

@inproceedings{kriz2025multivent2,
  title     = {{MultiVENT 2.0}: A Massive Multilingual Benchmark for Event-Centric Video Retrieval},
  author    = {Kriz, Reno and Sanders, Kate and Etter, David and Murray, Kenton and Carpenter, Cameron and Van Ochten, Kelly and Recknor, Hannah and Guallar-Blasco, Jimena and Martin, Alexander and Colaianni, Ronald and King, Nolan and Yang, Eugene and Van Durme, Benjamin},
  booktitle = {Proceedings of the IEEE/CVF Conference on Computer Vision and Pattern Recognition (CVPR)},
  year      = {2025}
}

@inproceedings{samuel2025mmmorrf,
  title     = {{MMMORRF}: Multimodal Multilingual Modularized Reciprocal Rank Fusion},
  author    = {Samuel, Saron and DeGenaro, Dan and Guallar-Blasco, Jimena and Sanders, Kate and Eisape, Oluwaseun and Spendlove, Tanner and Reddy, Arun and Martin, Alexander and Yates, Andrew and Yang, Eugene and Carpenter, Cameron and Etter, David and Kayi, Efsun and Wiesner, Matthew and Murray, Kenton and Kriz, Reno},
  booktitle = {Proceedings of the 48th International ACM SIGIR Conference on Research and Development in Information Retrieval (SIGIR)},
  year      = {2025}
}

@inproceedings{jeong2025videorag,
  title     = {{VideoRAG}: Retrieval-Augmented Generation over Video Corpus},
  author    = {Jeong, Soyeong and Kim, Kangsan and Baek, Jinheon and Hwang, Sung Ju},
  booktitle = {Findings of the Association for Computational Linguistics: ACL 2025},
  year      = {2025},
  pages     = {21278--21298}
}

@article{ren2025videorag,
  title   = {{VideoRAG}: Retrieval-Augmented Generation with Extreme Long-Context Videos},
  author  = {Ren, Xubin and Xu, Lingrui and Xia, Long and Wang, Shuaiqiang and Yin, Dawei and Huang, Chao},
  journal = {arXiv preprint arXiv:2502.01549},
  year    = {2025}
}

@article{zeng2025scenerag,
  title   = {{SceneRAG}: Scene-Level Retrieval-Augmented Generation for Video Understanding},
  author  = {Zeng, Nianbo and Hou, Haowen and Yu, Fei Richard and Shi, Si and He, Ying Tiffany},
  journal = {arXiv preprint arXiv:2506.07600},
  year    = {2025}
}

@article{sun2025f2c,
  title   = {From Frames to Clips: Efficient Key Clip Selection for Long-Form Video Understanding},
  author  = {Sun, Guangyu and Singhal, Archit and Uzkent, Burak and Shah, Mubarak and Chen, Chen and Kessler, Garin},
  journal = {arXiv preprint arXiv:2510.02262},
  year    = {2025}
}

@article{liang2024vidf4,
  title   = {End-to-End Video Question Answering with Frame Scoring Mechanisms and Adaptive Sampling},
  author  = {Liang, Jianxin and Meng, Xiaojun and Wang, Yueqian and Liu, Chang and Liu, Qun and Zhao, Dongyan},
  journal = {arXiv preprint arXiv:2407.15047},
  year    = {2024}
}

@article{zou2025air,
  title   = {{A.I.R.}: Enabling Adaptive, Iterative, and Reasoning-Based Frame Selection for Video Question Answering},
  author  = {Zou, Yuanhao and Liu, Yifan and Liu, Yang and Zhang, Yifan and Zhang, Han and Chen, Chen},
  journal = {arXiv preprint arXiv:2510.04428},
  year    = {2025}
}

@inproceedings{ye2025tstar,
  title     = {Re-Thinking Temporal Search for Long-Form Video Understanding},
  author    = {Ye, Jinhui and Wang, Zihan and Sun, Haosen and Chandrasegaran, Keshigeyan and Durante, Zane and Eyzaguirre, Cristobal and Bisk, Yonatan and Niebles, Juan Carlos and Adeli, Ehsan and Fei-Fei, Li and Wu, Jiajun and Li, Manling},
  booktitle = {Proceedings of the IEEE/CVF Conference on Computer Vision and Pattern Recognition (CVPR)},
  year      = {2025},
  pages     = {8579--8591}
}
\clearpage
\appendix

\twocolumn[
\begin{center}
    {\LARGE \bfseries Appendix}
\end{center}
\vspace{1em}
]
\section{MiRAGE Results using CLUE}
\Cref{tab:clue_reference_scores} reports per-query MiRAGE scores using CLUE as the evaluation backbone. The results show that CRAFT obtains stronger information precision than recall, indicating that its generated claims are often relevant and supported, but do not cover all reference subclaims. This is expected because CRAFT is designed to be conservative: it filters, deduplicates, and consolidates evidence to avoid unsupported statements, which improves factuality but can reduce coverage.

Citation scores are lower, especially citation recall. In MiRAGE, citation precision measures whether generated subclaims are supported by their cited videos, while citation recall measures whether the covered reference information is attributed to the correct supporting videos. This makes citation recall particularly challenging in MAGMaR, where evidence may be distributed across multiple heterogeneous videos and several videos may contain overlapping or partial support for the same event. As a result, a prediction can contain correct information but still lose citation recall if the exact supporting video is missing, incomplete, or not aligned with the evaluator's expected grounding. These results are therefore consistent with the main-text findings: CRAFT is relatively effective at producing factual content, but exact claim-to-video attribution remains the harder part of the task.



\begin{table}[t]
\centering
\tiny
\setlength{\tabcolsep}{3pt}
\renewcommand{\arraystretch}{1.12}
\resizebox{\columnwidth}{!}{%
\begin{tabular}{lcccc}
\toprule
\textbf{Topic} 
& \multicolumn{2}{c}{\textbf{Info F1}} 
& \multicolumn{2}{c}{\textbf{Cite F1}} \\
\cmidrule(lr){2-3} \cmidrule(lr){4-5}
& \textbf{P} & \textbf{R} & \textbf{P} & \textbf{R} \\
\midrule

\textbf{Average*} 
& \textbf{72.4} & \textbf{36.1} & \textbf{60.5} & \textbf{24.2} \\
\midrule

2025\_Myanmar\_earthquake\_q1 
& 72.7 & 86.7 & 59.1 & 80.0 \\

Liberation\_Day\_Tariffs\_q1 
& 70.0 & 64.1 & 65.0 & 66.7 \\

Blue\_Ghost\_Mission\_1\_q2 
& 70.8 & 57.1 & 70.8 & 39.3 \\

Shi\_Yongxin\_Scandal\_q1 
& 76.7 & 40.2 & 86.7 & 31.1 \\

Shi\_Yongxin\_Scandal\_q2 
& 95.2 & 36.4 & 95.2 & 30.3 \\

Blue\_Ghost\_Mission\_1\_q1 
& 76.5 & 39.3 & 64.7 & 35.7 \\

Liberation\_Day\_Tariffs\_q2 
& 70.6 & 41.0 & 70.6 & 30.8 \\

2025\_Alaskan\_Typhoon\_q2 
& 88.9 & 36.5 & 0.0 & 0.0 \\

Nepal\_Youth\_Protests\_q2 
& 92.9 & 29.4 & 96.4 & 23.5 \\

2025\_Alaskan\_Typhoon\_q1 
& 72.7 & 28.6 & 4.5 & 0.0 \\

Tropical\_Storm\_Wipha\_q1 
& 96.6 & 20.3 & 96.6 & 7.1 \\

2025\_Canadian\_federal\_election\_q2 
& 28.6 & 30.6 & 35.7 & 11.1 \\

Nepal\_Youth\_Protests\_q1 
& 63.6 & 13.2 & 63.6 & 4.4 \\

Palisades\_Fire\_q2 
& 100.0 & 9.5 & 100.0 & 2.6 \\

Palisades\_Fire\_q1 
& 78.3 & 7.9 & 47.8 & 2.1 \\

2025\_Canadian\_federal\_election\_q1 
& 3.6 & 36.1 & 10.7 & 22.2 \\

\bottomrule
\end{tabular}%
}
\vspace{2pt}
\caption{Per-topic CLUE reference scores for the CRAFT submission. Info F1 and Cite F1 are reported with precision (P) and recall (R). *We exclude queries with missing source videos from the MAGMaR-Test average, as these cases produce flat zero scores independent of system quality.}
\label{tab:clue_reference_scores}
\end{table}

\section{Human Evaluation}

The human evaluation, as shown in ~\cref{tab:human_eval_results}, indicate that CRAFT produces reasonably useful responses in several cases, but it is not yet consistently preferred over competing systems on MAGMaR leader-board. These results suggest that future improvements should focus on increasing information coverage and strengthening claim-to-video citation alignment, while preserving CRAFT's emphasis on grounded and conservative generation.

\begin{table}[t]
\centering
\tiny
\setlength{\tabcolsep}{3pt}
\renewcommand{\arraystretch}{1.12}
\resizebox{\columnwidth}{!}{%
\begin{tabular}{lccc}
\toprule
\textbf{Topic} & \textbf{Query} & \textbf{Avg. Score} & \textbf{Best Votes} \\
\midrule
\textbf{Overall} & -- & \textbf{2.542} & \textbf{0 / 57} \\
\midrule
2025-Alaska-typhoon & q1 & 3.000 & 0 / 3 \\
2025-Alaska-typhoon & q2 & 2.667 & 0 / 3 \\
2025\_Myanmar\_earthquake & q2 & 3.000 & 0 / 3 \\
2025\_Palisades\_fires & q1 & 2.667 & 0 / 3 \\
2025\_Palisades\_fires & q2 & 2.000 & 0 / 3 \\
2025\_Shi\_Yongxin\_Scandal & q1 & 3.000 & 0 / 3 \\
2025\_Shi\_Yongxin\_Scandal & q2 & 2.333 & 0 / 3 \\
2025\_Tropical\_Storm\_Wipha & q2 & 2.333 & 0 / 3 \\
2025\_canadian\_federal\_election & q1 & 2.000 & 0 / 3 \\
2025\_canadian\_federal\_election & q2 & 1.667 & 0 / 3 \\
2025\_nepal\_youth\_protests & q1 & 2.667 & 0 / 3 \\
2025\_nepal\_youth\_protests & q2 & 2.667 & 0 / 3 \\
Blue\_Mission\_Ghost\_1 & q1 & 2.333 & 0 / 3 \\
Blue\_Mission\_Ghost\_1 & q2 & 2.333 & 0 / 3 \\
Liberation-Day-tariffs & q1 & 3.333 & 0 / 3 \\
Liberation-Day-tariffs & q2 & 2.667 & 0 / 3 \\
\bottomrule
\end{tabular}%
}
\vspace{2pt}
\caption{Official human evaluation results for the CRAFT submission. Avg. Score is the mean scalar score on a 1--5 scale. Best Votes denotes the number of annotators who selected our system as the best response for the query. The overall scalar score is 2.542 with standard deviation 0.676 over 48 annotations.}
\label{tab:human_eval_results}
\end{table}

\section {Pre-Processing Details of WikiVideo}
To evaluate WikiVideo using the same structure as the MAGMaR test set, we convert the WikiVideo annotations into a MAGMaR-style format. We start with 56 candidate WikiVideo events and remove four events that overlap with the MAGMaR~2026 evaluation set, resulting in 52 events. For each event, we keep only reference claims that are supported by at least one video, and we further retain only events with at least three video-supported claims.

For each remaining event, an LLM agent generates a triplet <\texttt{persona\_title}, background, query> following the MAGMaR persona-query format. We then perform an audit step in which each generated triplet is scored on 5-point criteria on following axis: persona lignment, query answerability given article, and overall grounding. Items that are flagged during this audit are rewritten and rescored. The final dataset includes only events with an overall grounding score of at least 4, yielding a 52-query WikiVideo evaluation set.

Finally, the audited \texttt{persona\_title}, background, and query triples are converted into MAGMaR-style query, ground-truth, and topic-video files, allowing WikiVideo to be evaluated directly with the same pipeline used for MAGMaR.

\end{document}